# Updating with incomplete observations*

Gert de Cooman
Ghent University, SYSTeMS Research Group
Technologiepark – Zwijnaarde 914
9052 Zwijnaarde, Belgium
gert.decooman@ugent.be

Marco Zaffalon
IDSIA
Galleria 2, 6928 Manno (Lugano)
Switzerland
zaffalon@idsia.ch

## Abstract

Currently, there is renewed interest in the problem, raised by Shafer in 1985, of updating probabilities when observations are incomplete (or set-valued). This is a fundamental problem, and of particular interest for Bayesian networks. Recently, Grünwald and Halpern have shown that commonly used updating strategies fail here, except under very special assumptions. We propose a new rule for updating probabilities with incomplete observations. Our approach is deliberately conservative: we make no or weak assumptions about the so-called incompleteness mechanism that produces incomplete observations. We model our ignorance about this mechanism by a vacuous lower prevision, a tool from the theory of imprecise probabilities, and we derive a new updating rule using coherence arguments. In general, our rule produces lower posterior probabilities, as well as partially determinate decisions. This is a logical consequence of the ignorance about the incompleteness mechanism. We show how the new rule can properly address the apparent paradox in the 'Monty Hall' puzzle. In addition, we apply it to the classification of new evidence in Bayesian networks constructed using expert knowledge. We provide an exact algorithm for this task with linear-time complexity, also for multiply connected nets.

## 1 Introduction

This paper is concerned with the problem of updating probabilities with observations that are *incomplete*, or set-valued. To our knowledge, this problem was first given serious consideration in 1985 by Shafer [12]. He showed that it is at the heart of well-known puzzles, such as the Monty Hall puzzle. Moreover, his main argument was that the right way to update probabilities with incomplete observations requires knowledge of the *incompleteness mechanism* (called *protocol* in Shafer's paper), i.e., the mechanism that is responsible for turning a complete observation into an incomplete one, and he rightly observed that "we do not always have protocols in practical problems".

In practise, people often assume a condition known as *coarsening at random* (CAR [4]), or its specialisation to missing data problems, called *missing at random* (MAR [8]). These represent a form of partial knowledge about the incompleteness mechanism. Remarkably, when CAR holds, the common conditioning rule updates probabilities correctly. This may be one reason that Shafer's work seems to have been largely overlooked until 2002,[1] when an interesting paper by Grünwald and Halpern [5] offered a renewed perspective of the subject. This work argues strongly that CAR holds rather infrequently, and it enforces Shafer's viewpoint concerning the difficulties in knowing and modelling the incompleteness mechanism. These two points taken together raise a fundamental issue in probability theory, which also presents a serious problem for applications: how should beliefs be updated when there is little or no information about the incompleteness mechanism?

We believe that the first step is to allow for ignorance about the mechanism in our models. This is the approach that we take in this paper. In Section 3, we make our model as conservative as possible by representing the ignorance about the incompleteness mechanism by a *vacuous lower prevision*, a tool from the theory of imprecise probabilities[2] [14]. This theory is a generalisation of the Bayesian theory of probability [3], with a closely related behavioural interpretation, and based on similar criteria of rationality. Because we are aware that readers may not be familiar with imprecise probability models, we present a brief discus-sion in Section 2, with pointers to the relevant literature. Loosely speaking, the vacuous lower prevision is equivalent to the set of all distributions, i.e., it makes all incompleteness mechanisms possible *a priori*. Our basic model follows from this as a necessary consequence, using the rationality requirement of *coherence* (a generalisation of the requirements of rationality in Bayesian probability theory [3]). We illustrate how our basic model works by addressing the Monty Hall puzzle, showing that the apparent paradox vanishes if the available knowledge about the incompleteness mechanism is properly modelled.

---

*A longer version with proofs is available [2].

[1] But see the discussion by Walley in [14, Section 6.11], which has been a source of inspiration for the present work; and some papers by Halpern *et al.* [6, 7].

[2] See [15] for a gentle introduction to imprecise probabilities with emphasis on artificial intelligence.

We then apply our method for dealing with incomplete observations to the special case of a classification problem, where objects are assigned to classes on the basis of the values of their attributes. The question we deal with in Sections 4–6, is how classification should be done when values for some of the attributes are missing. We derive a new rule for updating, called *conservative updating rule*, that allows us to deal with such missing data without making unwarranted assumptions about the mechanism that produces them. Our rule leads to an imprecise posterior, and it may lead to inferences that are partially indeterminate. Our method will assign an object to a number of classes, rather than to a single class, unless conditions justify precision. This generalised way to do classification is called *credal classification* in [16]. Arguably this is the best our system can do, given the information that is incorporated into it. Also, any additional information about the missing data mechanism will lead to a new classification that refines ours, but can never contradict it by assigning an object to a class that was not among our optimal classes.

We then apply the conservative updating rule to classification problems with Bayesian networks. We regard a Bayesian net as a tool that formalises expert knowledge and is used to classify new evidence, i.e., to select certain values of a class variable given evidence about the attribute values. We develop an exact algorithm for credal classification with Bayesian nets that is linear in the number of children of the class node, also for multiply connected nets. This is an important result: it makes the new rule immediately available for applications; and it shows that it is possible for the power of robust, conservative, modelling to go hand in hand with easy computation, even with multiply connected networks, where the most common tasks are NP-hard.

## 2 Imprecise probabilities

The theory of coherent lower previsions (also called the theory of *imprecise probabilities*) [14] is an extension of the Bayesian theory of (precise) probability [3]. It models a subject's uncertainty by looking at his dispositions toward taking certain actions, and imposing requirements of rationality, or consistency, on these dispositions.

To make this more clear, consider a random variable $X$ that may take values in a finite set $\mathcal{X}$. A *gamble* $f$ on $\mathcal{X}$ is a real-valued function on $\mathcal{X}$. It associates a reward $f(x)$ with any possible value $x$ of $X$. If a subject is uncertain about the value of $X$, he will be disposed to accept certain gambles, and to reject others, and we may model his uncertainty by looking at which gambles he accepts (or rejects).

In the Bayesian theory of uncertainty (see for instance [3]), it is assumed that a subject can always specify a *fair price*, or *prevision*, $P(f)$ for $f$, whatever the information available to him. $P(f)$ is the unique real number such that he (i) accepts to buy the gamble $f$ for a price $p$, for all $p < P(f)$; and (ii) accepts to sell the gamble $f$ for a price $q$, for all $q > P(f)$. In other words, it is essentially assumed that for any real number $r$, the available information allows the subject to decide which of the following two options he prefers: buying $f$ for price $r$, or selling $f$ for that price.

It has been argued extensively [14] that, especially if little information is available about $X$, there may be prices $r$ for which a subject may have no real preference between these two options, or in other words, that on the basis of the available information he remains *undecided* about whether to buy $f$ or price $r$ or to sell it for that price: he may not be disposed to do either. If, as the Bayesian theory requires, the subject *should* choose between these two actions, his choice will then not be based on any real preference: it will be arbitrary, and not a realistic reflection of the subject's dispositions, based on the available information.

The theory of imprecise probabilities remedies this by allowing a subject to specify two numbers: $\underline{P}(f)$ and $\overline{P}(f)$. His *lower prevision* $\underline{P}(f)$ for $f$ is the greatest real number $p$ such that he is disposed to buy the gamble $f$ for all prices strictly smaller than $p$, and his *upper prevision* $\overline{P}(f)$ for $f$ is the smallest real number $q$ such that he is disposed to sell $f$ for all prices strictly greater than $q$. For any $r$ between $\underline{P}(f)$ and $\overline{P}(f)$, the subject does not express a preference between buying or selling $f$ for price $r$. Since selling a gamble $f$ for price $r$ is the same thing as buying $-f$ for price $-r$, we have the conjugacy relationship $\overline{P}(f) = -\underline{P}(-f)$ between lower and upper previsions. Whatever we say about upper previsions can always be reformulated in terms of lower previsions. We therefore concentrate on lower previsions. It suffices for our purposes to consider lower previsions $\underline{P}$ that are defined on the set $\mathcal{L}(\mathcal{X})$ of all gambles on $\mathcal{X}$, i.e., $\underline{P}$ is considered as a function that maps any gamble $f$ on $\mathcal{X}$ to the real number $\underline{P}(f)$.

An *event* $A$ is a subset of $\mathcal{X}$, and it will be identified with its *indicator function* $I_A$. We denote $\underline{P}(I_A)$ by $\underline{P}(A)$ and call it the *lower probability* of the event $A$. It is the supremum rate for which the subject is disposed to bet on the event $A$; and similarly for the *upper probability* $\overline{P}(A) = \overline{P}(I_A) = 1 - \underline{P}(\text{co}\, A)$. Thus, events are special gambles, and lower probabilities are special cases of lower previsions. We use the more general language of gambles, because Walley [14] has shown that in the context of imprecise probabilities, it is much more expressive and powerful.

Since lower previsions represent a subject's dispositions to act in certain ways, they should satisfy certain criteria that ensure that these dispositions are rational. *Coherence* is the strongest such requirement that is considered in the theory of imprecise probabilities. For a detailed definition and motivation, we refer to [14]. For our purposes, it suffices to expose the connections between coherent lower previsions and linear previsions, which are the coherent previsions in de Finetti's sense [3]. A *linear prevision* $P$ on $\mathcal{L}(\mathcal{X})$ is a real-valued map on $\mathcal{L}(\mathcal{X})$ satisfying the following properties: (i) $\min_{x \in \mathcal{X}} f(x) \leq P(f) \leq \max_{x \in \mathcal{X}} f(x)$; (ii) $P(f+g) = P(f)+P(g)$; and (iii) $P(\lambda f) = \lambda P(f)$; for all $f$ and $g$ in $\mathcal{L}(\mathcal{X})$, and all real numbers $\lambda$. Any linear prevision $P$ is completely determined by its so-called *mass function* $p$, defined by $p(x) = P(\{x\})$, since it follows from the axioms that for any gamble $f$, $P(f) = \sum_{x \in \mathcal{X}} f(x)p(x)$. We denote the set of all linear previsions on $\mathcal{L}(\mathcal{X})$ by $\mathcal{P}(\mathcal{X})$. Linear previsions are the so-called *precise* probability models, which will turn out to be special cases of the more general coherent imprecise probability models.

With any lower prevision $\underline{P}$ on $\mathcal{L}(\mathcal{X})$, we can associate its set of dominating linear previsions:

$$\mathcal{M}(\underline{P}) = \{P \in \mathcal{P}(\mathcal{X}) \colon (\forall f \in \mathcal{L}(\mathcal{X}))(\underline{P}(f) \leq P(f))\}.$$

It turns out that a lower prevision $\underline{P}$ on $\mathcal{L}(\mathcal{X})$ is *coherent* if and only if $\mathcal{M}(\underline{P}) \neq \emptyset$, and if moreover $\underline{P}$ is the lower envelope of $\mathcal{M}(\underline{P})$: for all gambles $f$ on $\mathcal{X}$, $\underline{P}(f) = \inf\{P(f) \colon P \in \mathcal{M}(\underline{P})\}$. Conversely, the lower envelope $\underline{P}$ of any non-empty subset $\mathcal{M}$ of $\mathcal{P}(\mathcal{X})$ is a coherent lower prevision. This tells us that working with coherent lower previsions is equivalent to working with sets of linear previsions. Observe that for a coherent $\underline{P}$, we have that $\overline{P}(f) \geq \underline{P}(f)$ for all $f \in \mathcal{L}(\mathcal{X})$.

There is a class of coherent lower previsions that deserves special attention. Consider a non-empty subset $B$ of $\mathcal{X}$. Then the *vacuous lower prevision* $\underline{P}_B$ *relative to* $B$ is defined by $\underline{P}_B(f) = \inf_{x \in B} f(x)$ for all gambles $f$ on $\mathcal{X}$. Verify that $\underline{P}_B$ is a coherent lower prevision, and moreover $\mathcal{M}(\underline{P}_B) = \{P \in \mathcal{P}(\mathcal{X}) \colon P(B) = 1\}$. This tells us that $\underline{P}_B$ is the smallest (most conservative) coherent lower prevision $\underline{P}$ on $\mathcal{L}(\mathcal{X})$ that satisfies $\underline{P}(B) = 1$. $\underline{P}(B) = 1$ means that it is *practically certain* to the subject that $X$ assumes a value in $B$, since he is prepared to bet at all odds on this event. Thus, $\underline{P}_B$ is the appropriate model for the piece of information that '$X$ assumes a value in $B$' *and nothing more*: any other coherent lower prevision $\underline{P}$ that satisfies $\underline{P}(B) = 1$ dominates $\underline{P}_B$, and therefore represents stronger behavioural dispositions than those required by coherence and this piece of information alone.

To introduce the concept of a conditional lower prevision, consider any gamble $h$ on $\mathcal{X}$ and any value $y$ in the finite set $\mathcal{Y}$ of possible values for another random variable $Y$. A subject's *conditional lower prevision* $\underline{P}(h|Y = y)$, also denoted as $\underline{P}(h|y)$, is his lower prevision for $h$ if he knew in addition that the variable $Y$ assumes the value $y$ (and nothing more!). We denote by $\underline{P}(h|Y)$ the gamble on $\mathcal{Y}$ that assumes the value $\underline{P}(h|Y = y) = \underline{P}(h|y)$ in $y \in \mathcal{Y}$. We can for the purposes of this paper assume that $\underline{P}(h|Y)$ is defined for all gambles $h$ on $\mathcal{X}$, and we call $\underline{P}(\cdot|Y)$ a conditional lower prevision on $\mathcal{L}(\mathcal{X})$. Observe that $\underline{P}(\cdot|Y)$ maps any gamble $h$ on $\mathcal{X}$ to the gamble $\underline{P}(h|Y)$ on $\mathcal{Y}$. Conditional lower previsions should of course also satisfy certain rationality criteria. $\underline{P}(\cdot|Y)$ is called *separately coherent* if for all $y \in \mathcal{Y}$, $\underline{P}(\cdot|y)$ is a coherent lower prevision on $\mathcal{L}(\mathcal{X})$. If besides the (separately coherent) conditional lower prevision $\underline{P}(\cdot|Y)$ on $\mathcal{L}(\mathcal{X})$, the subject has also specified a coherent (unconditional) lower prevision $\underline{P}$ on $\mathcal{L}(\mathcal{X} \times \mathcal{Y})$, then $\underline{P}$ and $\underline{P}(\cdot|Y)$ should in addition satisfy the consistency criterion of *joint coherence*. A discussion of this rationality requirement is beyond the scope of this paper, but we refer to [14, Chapter 6] for a detailed discussion and motivation.

There is an important and interesting procedure, called *regular extension*, that allows us to associate with any coherent lower prevision $\underline{P}$ on $\mathcal{L}(\mathcal{X} \times \mathcal{Y})$ a (separately coherent) conditional lower prevision $\underline{R}(\cdot|Y)$ that is jointly coherent with $\underline{P}$: (i) if $\overline{P}(\mathcal{X} \times \{y\}) = 0$, then $\underline{R}(\cdot|y)$ is the vacuous lower prevision relative to $\mathcal{X}$: $\underline{R}(h|y) = \inf_{x \in \mathcal{X}} h(x)$; and (ii) if $\overline{P}(\mathcal{X} \times \{y\}) > 0$, then $\underline{R}(h|y)$ is given by

$$\inf\left\{\frac{P(hI_{\mathcal{X} \times \{y\}})}{P(\mathcal{X} \times \{y\})} \colon P \in \mathcal{M}(\underline{P}), P(\mathcal{X} \times \{y\}) > 0\right\},$$

where $h$ is any gamble on $\mathcal{X}$. Thus, $\underline{R}(h|y)$ can be obtained by applying Bayes' rule (whenever possible) to all the precise previsions in $\mathcal{M}(\underline{P})$, and then taking the infimum! Regular extension has been called *divisive conditioning* by Seidenfeld *et al.* [11]. The regular extension is the smallest (most conservative) conditional lower prevision that is coherent with the joint $\underline{P}$ and satisfies an additional regularity condition [14, Appendix J].

We end this section with a discussion of decision-making using lower previsions. Suppose we have two actions $a$ and $b$, whose outcome depends on the actual value that the variable $X$ assumes in $\mathcal{X}$. Let us denote by $f_a$ the gamble on $\mathcal{X}$ representing the uncertain reward resulting from action $a$: a subject who takes action $a$ receives $f_a(x)$ if the value of $X$ turns out to be $x$. Similar remarks hold for $f_b$. If the subject is uncertain about the value of $X$, it is not immediately clear which of the two actions he should prefer, unless $f_a$ point-wise dominates $f_b$ or *vice versa*, which we shall assume is not the case. Suppose that he has modelled this uncertainty by a coherent lower prevision on $\mathcal{L}(\mathcal{X})$. Then he *strictly prefers* action $a$ to action $b$, which we denote as $a > b$, if he is willing to pay some strictly positive amount in order to exchange the (uncertain) rewards of $b$ for those of $a$. Using the behavioural definition of the lower prevision $\underline{P}$, this can be written as $a > b \Leftrightarrow \underline{P}(f_a - f_b) > 0$. If $\underline{P}$ is a linear prevision $P$, this is equivalent to $P(f_a) > P(f_b)$: the subject strictly prefers the action with the highest expected reward. It is easy to see that $\underline{P}(f_a - f_b) > 0$ can also be written as $(\forall P \in \mathcal{M}(\underline{P}))(P(f_a) > P(f_b))$. In other words, $a > b$ iff action $a$ yields a higher expected reward than $b$ for every linear prevision compatible with $\underline{P}$.

## 3 Incomplete observations

We are now ready to describe our basic model for dealing with incomplete observations. Consider a random variable $X$ that may assume values in a finite set $\mathcal{X}$. Suppose that we have some model for the available information about what value $X$ will assume in $\mathcal{X}$, which takes the form of a coherent lower prevision $\underline{P}_0$ defined on $\mathcal{L}(\mathcal{X})$. We now receive additional information about the value of $X$ by observing the value that another random variable $O$ assumes in a finite set $\mathcal{O}$. Only, these observations are *incomplete* in the following sense: the value of $O$ does not allow us to identify the value of $X$ uniquely. In fact, the only information we have about the relationship between $X$ and $O$ is the following: if we know that $X$ assumes the value $x$ in $\mathcal{X}$, then we know that $O$ must assume a value in a *non-empty* subset $\Gamma(x)$ of $\mathcal{O}$, *and nothing more*! This idea of modelling incomplete observations through a so-called *multi-valued map* $\Gamma$ essentially goes back to Strassen [13].

If we observe the value $o$ of $O$, then we know something more about $X$: it can then only assume values in the set $\{o\}^* = \{x \in \mathcal{X}: o \in \Gamma(x)\}$ of those values of $X$ that *may* produce the observation $O = o$. Unless $\{o\}^*$ is a singleton, the observation $O = o$ does not allow us to identify a unique value for $X$; it only allows us to restrict the possible values of $X$ to $\{o\}^*$. The question we want to answer here, is how we can use the new information that $O = o$ to coherently update the prior lower prevision $\underline{P}_0$ on $\mathcal{L}(\mathcal{X})$ to a posterior lower prevision $\underline{P}(\cdot|O = o) = \underline{P}(\cdot|o)$ on $\mathcal{L}(\mathcal{X})$.

In order to do this, we need to model the available information about the relationship between $X$ and $O$, i.e., about the so-called *incompleteness mechanism* that turns the values of $X$ into their incomplete observations $O$. In the special case that $\underline{P}_0$ is a (precise) linear prevision $P_0$ (with mass function $p_0$), it is often assumed that this mechanism obeys the CAR condition: $p(o|x) = p(o|y)$ for all $o \in \mathcal{O}$ and all $x$ and $y$ in $\{o\}^*$ such that $p_0(x) > 0$ and $p_0(y) > 0$ (see [4, 5] for an extensive discussion and detailed references). It is assumed that the probability of observing $O = o$ is not affected by the specific values $x$ of $X$ that may actually lead to this observation $o$. However, Grünwald and Halpern [5] have argued convincingly that CAR is a very strong assumption, that will only be justified in very special cases.

We want to refrain from making such unwarranted assumptions in general: we want to find out what can be said about the posterior $\underline{P}(\cdot|O)$ if *no* assumptions are made about the incompleteness mechanism, apart from those present in the definition of the multi-valued map $\Gamma$. This implies that anyone making additional assumptions (such as CAR) about the incompleteness mechanism will find results that are compatible but stronger, i.e., will find a posterior (lower) prevision that point-wise dominates ours.

We have argued in the previous section that the appropriate model for the piece of information that '$O$ assumes a value in $\Gamma(x)$' is the vacuous lower prevision $\underline{P}_{\Gamma(x)}$ on $\mathcal{L}(\mathcal{O})$ relative to the set $\Gamma(x)$. So we can model the relationship between $X$ and $O$ through the following (vacuous) conditional lower prevision $\underline{P}(\cdot|X)$ on $\mathcal{L}(\mathcal{O})$, defined by

$$\underline{P}(g|x) = \underline{P}_{\Gamma(x)}(g) = \inf_{o \in \Gamma(x)} g(o) \quad (1)$$

for any gamble $g$ on $\mathcal{O}$ and $x \in \mathcal{X}$. Using regular extension, we can now find the smallest (most conservative) conditional lower prevision $\underline{R}(\cdot|O)$ that is coherent with $\underline{P}_0$ and $\underline{P}(\cdot|\mathcal{X})$, and satisfies an additional regularity condition.

**Theorem 1.** *Let $o \in \mathcal{O}$ and let $f$ be any gamble on $\mathcal{X}$. If $\overline{P}(\{o\}^*) = 0$ then $\underline{R}(f|o) = \min_{x \in \mathcal{X}} f(x)$. If $\overline{P}_0(\{o\}^*) > 0$, then $\underline{R}(f|o)$ is the greatest value of $\mu$ such that, with $\{o\}_* = \{x \in \mathcal{X}: \Gamma(x) = \{o\}\}$,*

$$\underline{P}_0 \left(I_{\{o\}_*} \max\{f - \mu, 0\} + I_{\{o\}^*} \min\{f - \mu, 0\}\right) \geq 0.$$

Let us now apply the results of this theorem to the well-known Monty Hall puzzle. In the Monty Hall game show, there are three doors. One of them leads to a car, and the remaining doors each have a goat behind them. You indicate one door, and the show's host—let us call him Monty—now opens one of the other doors, which has a goat behind it. After this observation, should you choose to open the door that is left, rather than the one you indicated initially?

To solve the puzzle, we formulate it in our language of incomplete observations. Label the doors from 1 to 3, and assume without loss of generality that you picked door 1. Let the variable $X$ refer to the door hiding the car, then clearly $\mathcal{X} = \{1, 2, 3\}$. There is a precise prior prevision $P_0$ determined by $p_0(1) = p_0(2) = p_0(3) = 1/3$. The observation variable $O$ refers to the door Monty opens, and consequently $\mathcal{O} = \{2, 3\}$. If the car is behind door 1, Monty can choose between opening doors 2 and 3, so $\Gamma(1) = \{2, 3\}$, and similarly, $\Gamma(2) = \{3\}$ and $\Gamma(3) = \{2\}$. Since we don't know how Monty will choose between the options open to him, we should use Eq. (1) to model the available information about his choices (the incompleteness mechanism).

Assume without loss of generality that Monty opens door 2. If we apply Theorem 1, we find after some manipulations that $\underline{R}(f|2) = \frac{1}{2}f(3) + \frac{1}{2}\min\{f(3), f(1)\}$. Which of the two actions should we choose: stick to our initial choice and open door 1 (action $a$), or open door 3 instead (action $b$)? In Table 1 we see the possible outcomes of each action for the three possible values of $X$. If the gamble $f_a$ on $\mathcal{X}$ represents the uncertain reward received from action $a$, and similarly for $f_b$, then we are interested in the gamble $f_b - f_a$, which represents the uncertain reward from exchanging action $a$ for action $b$. This gamble is also in Table 1, where $\Delta > 0$ denotes the difference in utility between a car and a goat. Then we find that $\underline{R}(f_b - f_a|2) = 0$ and $\underline{R}(f_a - f_b|2) = -\Delta$. This implies that, with the notations established at the end of the previous section, $a \not\succ b$ and $b \not\succ a$: the available information does not allow us to say which of the two actions, sticking to door 1 (action $a$) or choosing door 3 (action $b$), is to be strictly preferred.

|   | 1 | 2 | 3 |
|---|---|---|---|
| $a$ | car | goat | goat |
| $b$ | goat | goat | car |
| $f_b - f_a$ | $-\Delta$ | 0 | $\Delta$ |

Table 1: Possible outcomes in the Monty hall puzzle

The same conclusion can also be reached as follows. Suppose first that Monty has decided on beforehand to always open door 3 when the car is behind door 1. Since he has actually opened door 2, the car cannot be behind door 1, and must therefore be behind door 3. In this case, action $b$ is clearly strictly preferable to action $a$. Next, suppose that Monty has decided on beforehand to always open door 2 when the car is behind door 1. Since he actually opens door 2, there are two equally likely possibilities, namely that the car is behind door 1 or behind door 3. Both actions now have the same expected reward (zero), and none is therefore strictly preferable to the other. Since both possibilities are consistent with the available information, we cannot infer any (robust) strict preference of one action over the other. A similar analysis was made by Halpern [6].

Observe that since $\underline{R}(f_b - f_a|2) = 0$, you *almost-prefer* $b$ to $a$, in the sense that you are disposed to exchange $f_a$ for $f_b$ in return for any strictly positive amount. In the case that Monty could also decide not to open any door, a similar analysis tells us that the updated lower prevision is given by $\underline{R}(f|2) = \min\{f(1), f(3)\}$, and we get $\underline{R}(f_b - f_a|2) = \underline{R}(f_a - f_b|2) = -\Delta$: neither option is even almost-preferred, let alone strictly preferred, over the other.

## 4 Missing data in a classification problem

In order to illustrate the practical implications of our model for the incompleteness mechanism, let us in the rest of this paper show how it can be applied in classification problems, where objects have to be assigned to a certain class on the basis of the values of their attributes.

Let in such a problem $\mathcal{C}$ be the set of possible classes that we want to assign objects to. Let $\mathcal{A}_1, \ldots, \mathcal{A}_n$ be the sets of possible values for the $n$ attributes on the basis of which we want the classify the objects. We denote their Cartesian product by $\mathcal{X} = \mathcal{A}_1 \times \cdots \times \mathcal{A}_n$. We consider a *class variable* $C$, which is a random variable in $\mathcal{C}$, and *attribute variables* $A_k$, which are random variables in $\mathcal{A}_k$ ($k = 1, \ldots, n$). The $n$-tuple $X = (A_1, \ldots, A_n)$ is a random variable in $\mathcal{X}$ and is called the *attributes variable*. The available information about the relationship between class and attribute variables is specified by a (prior) linear prevision $P_0$ on $\mathcal{L}(\mathcal{C} \times \mathcal{X})$.

Classification is done as follows: if the attributes variable $X$ assumes a value $x$ in $\mathcal{X}$, then the available information about the values of the class variable $C$ is clearly given by the conditional linear prevision $P_0(\cdot|x)$. If, on the basis of the observed value $x$ of the attributes variable $X$, we decide that some $c'$ in $\mathcal{C}$ is the right class, then we can see this classification as an action with an uncertain reward $f_{c'}$, whose value $f_{c'}(c)$ depends on the actual value $c$ of the class variable $C$. The discussion at the end of Section 2 then tells us that an optimal class $c_{opt}$ is one that maximises the expected reward $P_0(f_{c'}|x)$ over all $c' \in \mathcal{C}$. As an example, if we let $f_{c'} = I_{\{c'\}}$, then $P_0(f_{c'}|x) = p_0(c'|x)$, and this procedure associates the most probable class with each value $x$ of the attributes.

To make this more clear, let us consider a medical domain, where classification is employed to make a diagnosis. In this case, the classes are possible diseases and each attribute variable represents a measure with random outcome. For example, attribute variables might represent medical tests, or information about the patient, such as age, gender, life style, etc. We can regard the specific instance of the vector of attribute variables for a patient as a profile by which we characterise the person under examination. The relationship between diseases and profiles is given by a joint mass function on the class and the attribute variables. This induces a linear prevision $P_0$ on $\mathcal{L}(\mathcal{C} \times \mathcal{X})$, according to Section 2. A diagnosis is then obtained by choosing the most probable disease given a profile.

Now it may happen that for a patient some of the attribute variables cannot be measured, i.e., they are missing, as when for some reason a medical test cannot be done. In this case the profile is incomplete and we can regard it as the set of all the complete profiles that are consistent with it. The problem that we face is how we should update our confidence about the possible diseases given a set-profile, as the above classification procedure needs profiles to be complete.

In more general terms, we observe or measure the value $a_k$ of some of the attribute variables $A_k$, but not all of them. If a measurement is lacking for some attribute variable $A_\ell$, it can in principle assume any value in $\mathcal{A}_\ell$. We formalise this by associating with any attribute variable $A_k$ a so-called *observation variable* $O_k$. This is a random variable taking values in the set $O_k = A_k \cup \{*\}$, whose elements are either the possible values of $A_k$, or a new element $*$ which denotes that the measurement of $A_k$ is missing.

Attribute variables $A_k$ and their observations $O_k$ are linked in the following way: with each possible value $a_k \in A_k$ of $A_k$ there corresponds the following set of possible values for $O_k$: $\Gamma_k(a_k) = \{a_k, *\} \subseteq O_k$. This models that whatever value $a_k$ the attribute variable $A_k$ assumes, there is some mechanism, called the *missing data mechanism*, that either produces the (exact) observation $a_k$, or the observation $*$, which indicates that a value for $A_k$ is missing. For the attributes variable $X$ we then have that with each possible value $x = (a_1, \ldots, a_n)$ there corresponds a set $\Gamma(x) = \Gamma_1(a_1) \times \cdots \times \Gamma_n(a_n)$ of corresponding possible values for the *observations variable* $O = (O_1, \ldots, O_n)$, which assumes values in $\ddot{O} = \ddot{O}_1 \times \cdots \times \ddot{O}_n$.

So, in general, we observe some value $o = (o_1, \ldots, o_n)$ of the variable $O$, where $o_k$ is either the observed value for the $k$-th attribute, or $*$ if a value for this attribute is missing. In order to perform classification, we therefore need to calculate a coherent updated lower prevision $\underline{P}(\cdot|O = o)$ on $\mathcal{L}(\mathcal{C})$. This is what we now set out to do.

We have arrived at a special case of the model in the previous section, and the so-called missing data mechanism is a particular instance of the incompleteness mechanism described there. In this special case, it is easy to verify that the general CAR assumption, discussed previously, reduces to what is known in the literature as the MAR assumption [8]. MAR finds appropriate justification in some statistical applications, e.g., special types of survival analysis. However, there is strongly motivated criticism about the unjustified wide use of MAR in statistics, and there are well-developed methods based on much weaker assumptions [9].

As in the previous section, we want to refrain from making any strong assumptions about the mechanism that is behind the generation of missing values. We have argued before that the information in $\Gamma$, i.e., the information about the relationship between $X$ and $O$, can be represented by the conditional lower prevision $\underline{P}(\cdot|X)$ on $\mathcal{L}(\mathcal{O})$, defined by Eq. (1). We make the following additional *irrelevance assumption*: for all gambles $f$ on $\mathcal{C}$,

$$\underline{P}(f|x, o) = P_0(f|x) \text{ for all } x \in \mathcal{X} \text{ and } o \in \Gamma(x). \quad (I)$$

Assumption (I) states that, conditional on the attributes variable $X$, the observations variable $O$ is irrelevant to the class. In other words, the observations $o \in \Gamma(x)$ can influence our beliefs on the class only indirectly through the value $x$ of the attributes variable $X$, i.e., we are actually dealing with a problem of missing information. Note that the assumption is not restrictive in practise: if the fact that an attribute is missing can directly influence our beliefs on the class, then the related state $*$ should not be regarded as missing information, rather, as a possible value of the attribute, and it should be treated accordingly.

We can use regular extension to obtain the conditional lower prevision $\underline{R}(\cdot|O)$ on $\mathcal{L}(\mathcal{C})$. It is the smallest (most conservative) separately coherent conditional lower prevision that is jointly coherent with $P_0$ and $\underline{P}(\cdot|X)$, and takes into account the irrelevance assumption (I).[3]

**Theorem 2 (Conservative updating rule).** *Assume that irrelevance assumption* (I) *holds. Let $o$ be any element of $\mathcal{O}$. If $P_0(\{x\}) > 0$ for all $x \in \{o\}^*$, then $\underline{R}(f|o) = \min_{x:\ o \in \Gamma(x)} P_0(f|x)$ for all $f$ in $\mathcal{L}(\mathcal{C})$.*

Let us now denote by $E$ that part of the attributes variable $X$ that is instantiated, i.e., for which actual values are available. We denote its value by $e$. Let $R$ denote the other part, for whose components values are missing. We denote the set of its possible values by $\mathcal{R}$, and a generic element of that set by $r$. Then with some abuse of notation, $o = (e, *)$, and $\{o\}^* = \{e\} \times \mathcal{R}$. We then deduce from Theorem 2 that

$$\underline{R}(f|e, *) = \min_{r \in \mathcal{R}} P_0(f|e, r) \quad (2)$$

for all gambles $f$ on $\mathcal{C}$, provided that $p_0(e, r) > 0$ for all $r \in \mathcal{R}$, which we shall assume to be the case. We shall call Eq. (2) the *conservative updating rule*.

In the case of the earlier medical example, $e$ denotes the part of the profile that is known for a patient and the same incomplete profile can be regarded as the set $\{(e, r)|r \in \mathcal{R}\}$ of complete profiles that are consistent with it. The conservative updating rule tells us that in order to update our beliefs on the possible diseases given the incomplete profile, we have to consider all the complete profiles consistent with it, giving rise to lower and upper probabilities and previsions. In turn, this will generally give rise to partial classifications, according to the consideration about decision making in Section 2. That is, in general we will only be able to exclude some of the possible diseases given the evidence. This may give rise to a single disease, but only when the conditions justify precision.

The conservative updating rule is our main result: it provides us with the correct updating rule to use with an unknown incompleteness mechanism. It shows that robust, conservative, inference can be achieved by relying only on the original prior model of domain uncertainty.

## 5 Classification in expert systems with Bayesian networks

One popular way of doing classification in complex real-world domains involves using *Bayesian networks* (BNs). These are precise probabilistic models defined by a directed acyclic graph and a collection of conditional mass functions [10]. A generic node $Z$ in the graph is identified with a random variable. Each variable $Z$ holds a collection of conditional mass functions, one for each possible joint value $\pi_Z$ of its direct predecessor nodes (or *parents*) $\Pi_Z$. The generic conditional mass function assigns the probability $P_0(\{z\}|\pi_Z) = p_0(z|\pi_Z)$ to any value $z$ in $\mathcal{Z}$.

---

[3]And moreover satisfies an additional weak regularity condition, without which we would get completely vacuous inferences.

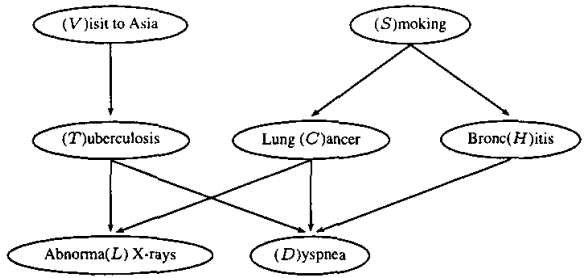

Figure 1: The 'Asia' Bayesian network.

Figure 1 displays the well-known example of a Bayesian network called 'Asia'.[4] It models an artificial medical problem by means of cause-effect relationships between random variables, e.g., $S \to C$ (each variable is denoted by the related letter between parentheses). The variables are binary and for any given variable, for instance $V$, its two possible values are denoted by $v'$ and $v''$, for the values 'yes' and 'no', respectively. The conditional probabilities for the variables of the model are reported in Table 2.

| $V = v'$ | 0.01 | | | | | | | |
|---|---|---|---|---|---|---|---|---|
| $S = s'$ | 0.5 | | | | | | | |
| | $v'$ | $v''$ | | | | | | |
| $T = t'$ | 0.05 | 0.01 | | | | | | |
| | $s'$ | $s''$ | | | | | | |
| $C = c'$ | 0.1 | 0.01 | | | | | | |
| | $s'$ | $s''$ | | | | | | |
| $H = h'$ | 0.6 | 0.3 | | | | | | |
| | $t'c'$ | $t'c''$ | $t''c'$ | $t''c''$ | | | | |
| $L = l'$ | 0.98 | 0.98 | 0.98 | 0.05 | | | | |
| | $t'c'h'$ | $t'c'h''$ | $t'c''h'$ | $t'c''h''$ | $t''c'h'$ | $t''c'h''$ | $t''c''h'$ | $t''c''h''$ |
| $D = d'$ | 0.9 | 0.7 | 0.9 | 0.7 | 0.9 | 0.7 | 0.8 | 0.1 |

Table 2: Asia example: probabilities for each variable in the graph conditional on the values of the parent variables.

Bayesian nets satisfy the *Markov condition*: every variable is stochastically independent of its non-descendant non-parents given its parents. Using the generic notation established in Section 4, assume the Bayesian network has nodes $C$, $A_1$, ..., $A_n$. From the Markov condition, it follows that the probability $p_0(c, a_1, \ldots, a_n)$ of a joint instance is given by $p_0(c|\pi_C) \prod_{i=1}^{n} p_0(a_i|\pi_{A_i})$, where the values of the parent variables are those consistent with $(c, a_1, \ldots, a_n)$. Hence, a BN is equivalent to a joint mass function over the variables of the graph. We assume that it assigns a strictly positive probability to any event.

Bayesian nets play an important role in the design of expert systems. Domain experts are supposed to provide both the qualitative graphical structure and the numerical values for the probabilities, thus implicitly defining an overall model of the prior uncertainty for the domain of interest. Users can then query the expert system, resulting in an update of the marginal prior probability of $C$ to a posterior probability according to the available evidence $E = e$, i.e., a set of nodes with known values. This kind of updating is very useful as it enables users to do classification, as we explain further on. In the Asia net, one might ask for the updated probability of lung cancer ($C = c'$), given that a patient is a smoker ($S = s'$) and has abnormal X-rays ($L = l'$), aiming ultimately at finding the proper diagnosis for the patient.

Updating the uncertainty for the class variable in a Bayesian net is subject to the considerations concerning incomplete observations in the preceding sections, as generally the evidence set $E$ will not contain all the attributes. To address this problem, one can assume that MAR holds and compute $p_0(c|e)$, but we have already pointed out that this approach is likely to be problematical in real applications.

Explicitly modelling the missing data mechanism is another way to cope with the problem, perhaps involving the same Bayesian net. The net would then also comprise the nodes $O_k$, $k = 1, \ldots, n$, for the observations; and the posterior probability of interest would become $p(c|o)$. Unfortunately, this approach presents serious practical difficulties. Modelling the mechanism can be as complex as modelling the prior uncertainty. Furthermore, it can be argued that in contrast with domain knowledge (e.g., medical knowledge), the way information can be accessed depends on the particular environment where a system will be used; and this means that models of the missing data mechanism will probably not be re-usable, and therefore costly.

These considerations support adopting a robust approach that can be effectively implemented, like the one we proposed in Section 4. We next develop an algorithm that exploits Eq. (2) to perform reliable classification with BNs.

## 6 An algorithm to classify incomplete evidence with BNs

How can we use the updated lower prevision $\underline{P}(\cdot|e, *)$ to perform classification? As in the case of a precise posterior $P_0(\cdot|x)$, we associate a reward function $I_{\{c'\}}$ with each class $c'$ in $\mathcal{C}$, and we look for those classes $c$ that are undominated elements of the strict partial order $>$ on $\mathcal{C}$, defined by (see Section 2)

$$c' > c'' \Leftrightarrow \underline{R}(I_{\{c'\}} - I_{\{c''\}}|e, *) > 0$$
$$\Leftrightarrow \min_{r \in \mathcal{R}} \frac{p_0(c', e, r)}{p_0(c'', e, r)} > 1, \quad (3)$$

where we have used Eq. (2), Bayes' rule, and the assumption that $p_0(e, r) > 0$ for all $r$ in $\mathcal{R}$. It is important to realise that the new updating will not always allow two classes to be compared, i.e., Eq. (2) generally produces only a partial order on the classes. As a consequence, the classification procedure consists in comparing each pair of classes by strict preference (also called *credal dominance* in [16]) and in discarding the dominated ones. The system will then output a set of *possible*, undominated classes. Classifiers with this characteristic are also called *credal classifiers*. In the following we address the efficient computation of the credal dominance test (3).

---

[4]The network presented here is equivalent to the traditional one although it is missing a logical OR node.

Let $\pi'$ and $\pi''$ denote values of the parent variables consistent with $(c', e, r)$ and $(c'', e, r)$, respectively. If a node's parents do not contain $C$, let $\pi$ denote the value of the parent variables consistent with $(e, r)$. Furthermore, without loss of generality, let $A_1, \ldots, A_m$, $m \leq n$, be the children of $C$, and $K = \{1, \ldots, m\}$. Let $B$ be the *Markov blanket* of $C$, that is, the set of nodes consisting of the parents of $C$, its children, and the parents of the children of $C$.

Consider a total order on the children of $C$ that extends the partial order given by the arcs of the graph, i.e., when $A_i \to A_j$ is interpreted as: $A_i$ precedes $A_j$ in the partial order. By permuting the subscripts we can always say that the total order is just $A_1 \to A_2 \to \cdots \to A_m$. Let $A_0 = C$. For each $i = 0, \ldots, m$, let $\Pi_{A_i}^+ = \Pi_{A_i} \cup \{A_i\}$, and for $i \in K$ let $\Pi_{A_i}^* = \bigcup_{j=0}^{i-1} \Pi_{A_i}^+$. Now let

$$\mu_{A_{m+1}} = 1$$

$$\mu_{A_i} = \min_{\substack{a_j \in \mathcal{A}_j \\ A_j \in (\Pi_{A_i}^+ \setminus \Pi_{A_i}^*) \cap R}} \left[ \frac{p_0(a_i | \pi'_{A_i})}{p_0(a_i | \pi''_{A_i})} \mu_{A_{i+1}} \right], \forall i \in K$$

$$\mu_{A_0} = \min_{\substack{a_j \in \mathcal{A}_j \\ A_j \in \Pi_C \cap R}} \left[ \frac{p_0(c' | \pi_C)}{p_0(c'' | \pi_C)} \mu_{A_1} \right] \quad (4)$$

Note that $\mu_{A_i}$, $i = 1, \ldots, m$, are not constants in general but functions of some attribute variables, as the minimisation does not always involve all the attributes of which the argument of $\mu_{A_i}$ is function.

**Theorem 3.** *Consider a BN with nodes $C$, $A_1$, ..., $A_n$. Let $c', c'' \in \mathcal{C}$. Then $c'$ credal-dominates $c''$ if and only if $\mu_{A_0} > 1$, where $\mu_{A_0}$ is the optimal solution of Eq. (4).*

Theorem 3 renders the solution of the credal-dominance test very easy, in a dynamic programming fashion. One starts by computing the last minimum $\mu_{A_m}$. The produced function is multiplied by the preceding ratio and the new function $\mu_{A_{m-1}}$ is evaluated. The process continues until $\mu_{A_0}$ (i.e., $\mu_C$) is computed. The overall computational complexity is $O(m+1)$.

As an example, consider the Asia net, where we choose $C$ as the class. Its Markov blanket of $C$ is $\{S, L, D, T, H\}$. The graph presents no relationship of order between $L$ and $D$, and we arbitrarily choose $L \to D$. Then we set the evidence to $L = l'$ and $S = s'$.

We want to test whether $c'$ credal-dominates $c''$. We start by computing

$$\mu_D(t) = \min_{d \in \mathcal{D}, h \in \mathcal{H}} \frac{p_0(d|t, c', h)}{p_0(d|t, c'', h)}.$$

We have $\mu_D(t') = \min\left\{\frac{0.9}{0.9}, \frac{0.7}{0.7}, \frac{0.1}{0.1}, \frac{0.3}{0.3}\right\} = 1$ and $\mu_D(t'') = \min\left\{\frac{0.9}{0.8}, \frac{0.7}{0.1}, \frac{0.1}{0.2}, \frac{0.3}{0.9}\right\} = \frac{1}{3}$. Next, we compute

$$\mu_L = \min_{t \in \mathcal{T}} \left[ \frac{p_0(l'|t, c')}{p_0(l'|t, c'')} \mu_D(t) \right].$$

We have $\mu_L = \min\left\{\frac{0.98}{0.98} 1, \frac{0.98}{0.05} \frac{1}{3}\right\} = 1$. Finally, we compute

$$\mu_C = \mu_L \frac{p_0(c'|s')}{p_0(c''|s')} = 1 \frac{0.1}{0.9} = \frac{1}{9}.$$

As $\mu_C$ is not greater than 1, $c''$ is undominated.

Testing whether $c''$ credal-dominates $c'$ is very similar and leads to $\mu_C = \frac{45}{686}$, so $c'$ is undominated as well. In this situation, the system suspends judgement, i.e., it outputs both the classes, as there is not enough information to allow us to choose between the two. This should be contrasted with traditional updating, which produces $p_0(c'|l', s') \simeq 0.646$, and leads us to diagnose cancer.

It is useful to better analyse the reasons for the indeterminate output of the proposed system. Given our assumptions, the system cannot exclude that the available evidence, or incomplete profile, is part of a more complete profile where $T = t'$, $D = d'$, and $H = h'$. If this were the case, then $c''$ would be nine times as probable *a posteriori* as $c'$, and we should diagnose no cancer. However, the system cannot exclude either that the more complete profile would be $T = t''$, $D = d'$, and $H = h''$. In this case, the ratio of the posterior probability of $c'$ to that of $c''$ would be $\frac{686}{45}$, leading us to the opposite diagnosis.

Of course when the evidence is strong enough, the proposed system does produce determinate conclusions. For instance, the incomplete profile given by $L = l'$, $S = s'$ and $T = t'$, will lead the system to exclude the presence of cancer.

## 7　Conclusions

We have proposed a conservative rule for updating probabilities with incomplete observations when strong assumptions about the incompleteness mechanism cannot be justified, thus filling an important gap in literature. We have achieved this result by coherent lower previsions, following an approach very similar in spirit to the Bayesian method. However, imprecise probabilities allowed us to work naturally also with generalised priors, such as the vacuous prior, which are needed to model states of partial or total ignorance. Generalised priors are significantly more expressive than traditional priors: they enable a new set of difficult important problems to be addressed in an adequate and elegant manner, as this paper shows in an enlightening case.

By focusing on classification of new evidence we have shown that the conservative updating leads to a very efficient implementation, also for multiply connected nets, so the new developments can immediately be exploited in real environments. Furthermore, the related algorithm can be implemented easily and does not require changes in pre-existing knowledge bases, so that existing expert systems can be upgraded to make our robust, conservative, inferences with minimal changes.

The proposed updating strategy is different in one important respect from the more traditional ones: it generally leads only to partially determinate inferences and decisions, and ultimately to systems that can recognise the limits of their knowledge, and suspend judgement when these limits are reached. As necessary consequences of our refusal to make unwarranted assumptions, we believe that these limitations are important characteristics of the way systems ought to operate in the real world. A system that, in a certain state, cannot support any decision on the basis of its knowledge base, will induce a user to look for further sources of information externally to the system. In contrast, systems that may make arbitrary choices without making that evident, will wrongly lead a user to think that these choices are well motivated.

An important subject for future research is the identification of common intermediate states of knowledge about the incompleteness mechanism, that can be usefully modelled to derive stronger inferences and decisions than the ones described here. For Bayesian nets, one could think of partitioning the set of attributes in those for which MAR holds and the rest for which the mechanism is unknown. Such hybrid modelling seems a good compromise between generality and flexibility. It is also very useful to consider the extension of our treatment for Bayesian networks to the more general notion of a *credal network* [1]. Credal nets extend Bayesian nets in that they allow for imprecise probability models (sets of probability distributions). They permit much more flexible modelling by weakening the requirement that prior knowledge should be represented by a precise probability distribution. Recent investigations make us confident about the extension of the present work to credal networks.

### Acknowledgements

The authors are grateful to Peter Walley for stimulating discussions on the topic of the paper in August 2001. This research was partially supported by the Swiss NSF grant 2100-067961.02, and by research grant G.0139.01 of the Flemish Fund for Scientific Research (FWO).

### References

[1] F. G. Cozman. Credal networks. *Artificial Intelligence*, 120:199–233, 2000.

[2] G. de Cooman and M. Zaffalon. Updating probabilities with incomplete observations. Technical Report IDSIA-12-03, IDSIA, 2003.

[3] B. de Finetti. *Theory of Probability*, volume 1. John Wiley & Sons, Chichester, 1974. English Translation of *Teoria delle Probabilità*.

[4] R. Gill, M. Van der Laan, and J. Robins. Coarsening at random: characterisations, conjectures and counter-examples. In D.-Y. Lin, editor, *Proceedings of the first Seattle Conference on Biostatistics*, pages 255–294. Springer, 1997.

[5] P. Grünwald and J. Halpern. Updating probabilities. In A. Darwiche and N. Friedman, editors, *Proceedings of the 18th Conference on Uncertainty in Artificial Intelligence (UAI-2002)*, pages 187–196. Morgan Kaufmann, 2002.

[6] J. Y. Halpern. A logical approach to reasoning about uncertainty: a tutorial. In X. Arrazola, K. Korta, and F. J. Pelletier, editors, *Discourse, Interaction, and Communication*, pages 141–155. Kluwer, 1998.

[7] J. Y. Halpern and M. Tuttle. Knowledge, probability, and adversaries. *Journal of the ACM*, 40(4):917–962, 1993.

[8] R. J. A. Little and D. B. Rubin. *Statistical Analysis with Missing Data*. Wiley, New York, 1987.

[9] C. Manski. *Partial Identification of Probability Distributions*. Springer, 2003. Forthcoming.

[10] J. Pearl. *Probabilistic Reasoning in Intelligent Systems: Networks of Plausible Inference*. Morgan Kaufmann, San Mateo, CA, 1988.

[11] T. Seidenfeld and L. Wasserman. Dilation for sets of probabilities. *The Annals of Statictics*, 21:1139–54, 1993.

[12] G. Shafer. Conditional probability. *International Statistical Review*, 53:261–277, 1985.

[13] V. Strassen. Meßfehler und Information. *Zeitschrift für Wahrscheinlichkeitstheorie und Verwandte Gebiete*, 2:273–305, 1964.

[14] P. Walley. *Statistical Reasoning with Imprecise Probabilities*. Chapman and Hall, London, 1991.

[15] P. Walley. Measures of uncertainty in expert systems. *Artificial Intelligence*, 83:1–58, 1996.

[16] M. Zaffalon. The naive credal classifier. *Journal of Statistical Planning and Inference*, 105:5–21, 2002.